\documentclass[preprint]{article}

 \usepackage{neurips_2025}

\usepackage[utf8]{inputenc} 
\usepackage[T1]{fontenc}    
\usepackage{hyperref}       
\usepackage{url}            
\usepackage{booktabs}       
\usepackage{amsfonts}       
\usepackage{nicefrac}       
\usepackage{microtype}      
\usepackage{xcolor}         
\usepackage{multirow}
\usepackage{graphicx}
\usepackage{adjustbox}
\usepackage{booktabs}
\usepackage{float}
\usepackage{xcolor,colortbl}
\usepackage{makecell}
\usepackage{amsmath}
\usepackage{minted}
\usepackage{subcaption}
\title{Brittleness and Promise: Knowledge Graph–Based Reward Modeling for Diagnostic Reasoning}

\author{
  Saksham Khatwani \\
  University of Colorado Boulder \\
  Boulder, CO \\
  \texttt{saksham.khatwani@colorado.edu} \\
  \And
  He Cheng \\
  University of Colorado Anschutz \\
  Aurora, CO\\
  \texttt{he.2.cheng@cuanschutz.edu} \\
  \And
  Majid Afshar \\
  University of Wisconsin - Madison\\
  Madison, WI \\
  \texttt{mafshar@medicine.wisc.edu} \\
  \And
  Dmitriy Dligach \\
  Loyola University \\
  Chicago, IL \\
  \texttt{ddligach@luc.edu} \\
  \And
  Yanjun Gao \\
  University of Colorado Anschutz \\
  Aurora, CO \\
  \texttt{yanjun.gao@cuanschutz.edu} \\
}

\begin{document}

\maketitle

\begin{abstract}
    Large language models (LLMs) show promise for diagnostic reasoning but often lack reliable, knowledge-grounded inference. Knowledge graphs (KGs), such as the Unified Medical Language System (UMLS), offer structured biomedical knowledge that can support trustworthy reasoning. Prior approaches typically integrate KGs via retrieval-augmented generation or fine-tuning, inserting KG content into prompts rather than enabling structured reasoning. We explore an alternative paradigm: treating the LLM as a \textit{reward} model of KG reasoning paths, where the model learns to \textit{judge} whether a candidate path leads to correct diagnosis for a given patient input. This approach is inspired by recent work that leverages reward training to enhance model reasoning abilities, and grounded in computational theory, which suggests that \textit{verifying} a solution is often easier than generating one from scratch. It also parallels physicians’ diagnostic assessment, where they \textit{judge} which sequences of findings and intermediate conditions most plausibly support a diagnosis. We first systematically evaluate five task formulation for knowledge path judging and eight training paradigm. Second, we test whether the path judging abilities generalize to downstream diagnostic tasks, including diagnosis summarization and medical question answering. Experiments with three open-source instruct-tuned LLMs reveal both promise and brittleness: while specific reward optimization and distillation lead to strong path-judging performance, the transferability to downstream tasks remain weak. Our finding provides the first systematic assessment of ``reward model style'' reasoning over clinical KGs, offering insights into how structured, reward-based supervision influences diagnostic reasoning in GenAI systems for healthcare.    
    
\end{abstract}

\section{Introduction}

Large language models (LLMs) have shown remarkable potential in supporting diagnostic reasoning, enabling them to generate differential diagnoses, summarize clinical information, and answer complex medical questions~\cite{goh2024large,wang2025medical,liu2025generalist}. Yet, despite their fluency, LLMs frequently fall short in terms of faithful reasoning: they may overlook relevant evidence, hallucinate unsupported conclusions, or rely on superficial statistical associations rather than structured medical knowledge~\cite{kim2025medical,tonmoy2024comprehensive,asgari2025framework}. For deployment in clinical practice, these limitations highlight the urgent need to augment LLMs with knowledge-grounded reasoning mechanisms.

Knowledge graphs (KGs) offer a promising pathway toward trustworthy diagnostic reasoning. Richly structured resources such as the Unified Medical Language System (UMLS)~\cite{Bodenreider2004TheUM} encode decades of curated biomedical concepts and relations, serving as a foundation for aligning free-text patient descriptions with symbolic diagnostic pathways. Recent efforts demonstrate the potential of combining KGs with LLMs to improve diagnostic generation, such as \textsc{DR.Knows}~\cite{gao2025leveraging}, \textsc{KG4Diagnosis}~\cite{zuo2025kg4diagnosis}, and other approaches like prompting with graph-of-thoughts~\cite{bedi2025xlr,wen2024mindmap}. These studies suggest that neural–symbolic integration can increase both the accuracy and interpretability of diagnostic reasoning.

Two primary strategies have emerged for incorporating KGs into LLMs. The first is graph-based retrieval-augmented generation (RAG), where an external retriever identifies relevant KG subgraphs or paths and inserts them into the LLM’s prompt as additional context, as exemplified in~\cite{gao2025leveraging,wen2024mindmap,chen2024new}. The second is to fine-tune LLMs in KG-enriched corpora, so that the model implicitly learns the graph structure during parameter updates~\cite{wang2025medical, tian-etal-2024-kg,chen2025knowledge,chen2024new}. While effective to some extent, both approaches face challenges: RAG suffers from incomplete or noisy retrieval, while fine-tuning is costly and often fails to generalize across domains or tasks.

Recent work has introduced the paradigm of the ``reward model as reasoning'': showing that reward models can go beyond scalar scoring to perform meaningful reasoning during evaluation, yielding better downstream performance than traditional reward modeling approaches~\cite{chen2025reasongrm,chen2025rm}. We extend this idea to the medical KG settings, where the task is not to generate a diagnostic path but to \textit{judge whether a candidate path is clinically valid for a given patient}. This framing is motivated by a simple but powerful observation: verification is often easier than solution. In computational theory, many problems are hard to solve but straightforward to verify~\cite{cook2023complexity,sipser2012introduction,godel1956letter,swamy2025all}. Likewise, searching through a large, sparse KG such as UMLS to identify the correct diagnostic path is challenging. In contrast, assessing whether a given path is clinically relevant to the patient’s condition, while still requiring reasoning, can be more tractable than generating the path from scratch. 

In our setting, we define \textit{reward-model-as-reasoning} as training an LLM to assess whether a KG-derived path meaningfully connects a patient’s clinical findings to their diagnoses. This goes beyond lexical or semantic similarity: the model must reason over potentially overlapping or competing concepts (e.g., “Type 1” vs. “Type 2” diabetes), determine which paths reflect appropriate causal or diagnostic mechanisms, and weigh relevance based on the patient’s condition. In short, the reward model must emulate how a clinician might evaluate whether a given explanation fits the clinical picture. We explore this paradigm using the Unified Medical Language System (UMLS), one of the most widely adopted KGs in clinical informatics, maintained by the U.S. National Library of Medicine for over three decades~\cite{Bodenreider2004TheUM}. While our study is limited in scale, we take a systematic approach to investigate: (i) which task formulations and training methods best support reward-style reasoning over KG paths, and (ii) whether such capabilities transfer to broader diagnostic reasoning tasks such as diagnosis summarization and medical question answering. 

\section{Related Work}
\paragraph{KG–LLM Integration for Diagnosis} 
An increasing number of studies integrate KGs with LLMs to support diagnostic predictions. \citet{Gao_2025} propose \textit{DR.KNOWS}, which retrieves relevant paths based on patient-specific information from UMLS KGs to improve LLMs' diagnostic predictions. \citet{zuo2025kg4diagnosis} present an end-to-end hierarchical multi-agent framework, \textit{KG4Diagnos}, where one primary diagnostic agent processes user queries and multiple specialized diagnostic agents collaborate to improve diagnostic predictions. \citet{jia2024medikal} develop \textit{medIKAL}, in which KG knowledge assists LLMs in generating and ranking candidate diagnoses, leading to more robust performance on EMR-based tasks. \citet{xie2025kerap} introduce \textit{KERAP}, a multi-agent pipeline that links patient information to KG entities, retrieves relevant knowledge, and predicts diagnoses in a zero-shot setting. \citet{zhao2025medrag} present \textit{MedRAG}, which combines RAG with a structured diagnostic KG to refine EHR-based predictions. In general, these systems demonstrate that KG integration can make diagnostic LLMs more accurate, interpretable, and clinically useful.

\paragraph{Reward Models as Reasoning} 
Recent work views reward modeling as a reasoning task and moves beyond simple preference scores toward models that check the correctness of a solution. \citet{chen2025rm} frame reward modeling as a reasoning task by distilling high-quality traces and combining them with reinforcement learning and structured Chain-of-Rubrics rollouts. \citet{guo2025reward} propose \textit{RRMs} that perform explicit reasoning at inference time before scoring, training without reasoning traces but learning to use extra test-time compute on difficult cases. \citet{chen2025reasongrm} propose \textit{ReasonGRM}, which improves reward models by generating outcome-aligned reasoning traces, filtering them with a new $R^{\star}$ metric, and refining with RL on hard cases. \citet{xu2025direct} introduce \textit{DRO}, which defines a reasoning reflection reward that aligns a model’s chain-of-thought with its final outcome by emphasizing reasoning-reflective tokens. In summary, reward models as reasoning go beyond scalar outcome evaluation by treating the reward model itself as a reasoning agent that generates, critiques, and aligns reasoning traces to produce more interpretable and reliable judgments.

\paragraph{LLM Graph Reasoning Techniques} 
Recent work augments LLMs with explicit graph structures, moving beyond linear chain-of-thought prompting toward more faithful and interpretable reasoning. \citet{luo2024reasoning} introduce RoG, which grounds reasoning in KG relation paths through a planning–retrieval–reasoning framework. \citet{jin2024graph} present Graph-CoT, which improves LLM reasoning by alternating between generating reasoning steps and interacting with text-attributed graphs. \citet{han2025reasoning} propose \textit{RwG}, which builds explicit graphs from context through iterative generation and verification, enabling reasoning over structured knowledge. \citet{hu2024scalable} propose \textit{GAR}, a multi-agent framework where graph nodes act as LLM agents exchanging messages, coordinated by a Master LLM through a message-passing scheme. Together, these works mark a shift from text-only prompting to graph-based methods that structure reasoning through planning, traversal, and distributed computation.

\paragraph{Our Contribution}

Building on the aforementioned work, we explore a new direction:  whether fine-tuning an LLM to reason diagnostically by learning to assess the validity of a KG knowledge path. While prior work has focused on retrieval systems, symbolic pipelines, or reward models for evaluation, our work is the \textit{first to apply} reward style path supervision directly for training diagnostic language models. 


\section{Data Overview}
\subsection{Dataset}
\paragraph{UMLS Knowledge Graph}
UMLS is a large-scale resource that integrates a wide range of biomedical vocabularies and standards, maintained by U.S. National Library of Medicine over three decades. It provides comprehensive concept vocabularies from 180 biomedical sources, and defines 270 semantic relationships, which enable the construction of KGs through concepts and relations. In this work, we leverage the KG version introduced in \cite{Gao_2025}, which is constructed specifically for diagnostic reasoning with physician-selected relations and concept vocabularies. This KG includes 107 relations most pertinent to diagnostic reasoning, and concepts specifically from the Systematized Nomenclature of Medicine – Clinical Terms (SNOMED CT) vocabulary. Each concept in the UMLS knowledge graph is assigned a SNOMED CT concept unique identifier (CUI) and categorized by semantic type, enabling multi-hop exploration across the graph. 

\paragraph{ProbSum Dataset}
We employed the dataset introduced in the BioNLP 2023 ProbSum shared task \cite{gao-etal-2023-overview}, which consists of daily patient progress notes paired with physician annotated diagnoses. These notes were collected from MIMIC-III~\cite{Johnson2016MIMICIIIAF}, representing a cohort of ICU patients who on average have 5 active diagnoses~\cite{gao-etal-2023-overview}. The training set comprises 768 progress notes covering 2,783 diagnoses, while the test set contains 237 progress notes.

\paragraph{MedQA}
The MedQA dataset~\cite{jin2020diseasedoespatienthave} contains multiple-choice medical questions with four answer options and one correct answer. It spans various clinical scenarios, including treatment, management, and diagnosis. For our study, we focused on diagnosis-related questions, yielding 1,796 training and 251 test samples. These were used to fine-tune and evaluate our models in a different diagnostic reasoning setting.

\subsection{Building supervision paths for diagnostic reasoning}
To construct diagnosis-oriented reasoning paths, we leverage both patient progress notes and corresponding diagnosis labels from the ProbSum dataset. For each patient, we first apply QuickUMLS \cite{Soldaini2016QuickUMLSAF} to extract medical concepts from the clinical note, which serve as candidate starting points in the UMLS knowledge graph. In parallel, we identify the patient’s gold-standard diagnoses, also mapped to UMLS concepts. While UMLS concepts span a wide range of semantic types, not all are directly relevant to diagnosis. Accordingly, we restricted our analysis to the following categories: T033 (Finding), T037 (Injury or Poisoning), T046 (Pathologic Function), T047 (Disease or Syndrome), T048 (Mental or Behavioral Dysfunction), T049 (Cell or Molecular Dysfunction), and T184 (Sign or Symptom).

From each starting concept node, we performed a depth-first search (DFS) over the UMLS KG, exploring all connected nodes with maximum two-hop. If a path terminates at the gold-standard diagnosis concept, it is labeled as a positive path; otherwise, it is treated as a negative path. This procedure yields multiple path samples (positive and negative) per patient note.

To create training data, we randomly select 326 patient notes from the 768 in the ProbSum training set. Across these notes, we collect 13,377 unique starting concepts, resulting in 27,535 path-based training examples (average of 84 per note). These examples are then formatted into two task-specific datasets:
\begin{itemize}
    \item In the path selection task, the model receives a progress note and a mixed set of candidate paths (positive and negative) and must identify the valid ones. This requires judgment, comparison, and an understanding of global coherence in the KG. It’s akin to reviewing multiple differential diagnoses and determining which fit the patient context. 
    \item In the path completion task, the model is shown a progress note and a partially completed positive path and must predict the missing step. This requires the model to simulate plausible causal or diagnostic chains, leveraging local graph structure, relation semantics, and disease progression patterns. 
\end{itemize}

\paragraph{Evaluation} All models are trained on the same pool of 27,535 samples, formatted according to the respective task setup. Because our framework follows a generative AI paradigm, we employ two widely used generative evaluation metrics: ROUGE-1 and ROUGE-L~\cite{lin2004rouge}. These are used for all path-specific judging tasks. For downstream evaluation, we adopt the metric convention of each benchmark: ROUGE-L for the ProbSum summarization task, and exact match accuracy for MedQA, which is framed as multiple-choice question answering.In addition to conventional metrics, we adopt PDSQI-9~\cite{croxford2025developmentvalidationproviderdocumentation}, a clinically validated rubric for LLM-generated summaries. It includes nine criteria covering abstraction accuracy, evidence use, thoroughness, and organization. Following prior work, we use Azure GPT-4o-mini (ICC $\geq$ 0.8) as the LLM-as-judge to score the ProbSum test set. The Azure is a HIPPA-Complaint environment therefore obey the MIMIC Data Use Agreement.

\section{Methods} \label{Methods}

\subsection{Task formulation}
To teach and evaluate whether LLMs can reason over medical knowledge graphs in support of diagnosis, we design five task settings corresponding to the two setup mentioned above: 

\begin{itemize}
    \item \textsc{\textbf{P@10}-Path selection}: The model is given 10 candidate KG paths, exactly one of which is valid.
    \item \textsc{\textbf{P@2}-Path selection}: The model is given 2 candidate KG paths, one valid and one invalid.
    \item \textsc{\textbf{PN@10}-Path selection}: The model is given 10 candidate KG paths, of which multiple may be valid, and is required to identify all correct paths.
    \item \textsc{\textbf{Next-Hop Prediction (NHP)}-Path completion}: Given a partial KG path, the model must predict the next hop.
    \item \textsc{\textbf{Path Completion (PC)}}: Given a KG partial path, the model must predict the remaining sequence of hops to complete the path.
    
\end{itemize}

We hypothesize that these task settings expose complementary aspects of reasoning over the KGs. The \textsc{P@10}, \textsc{P@2}, and \textsc{PN@10} settings emphasize the ability to evaluate entire candidate paths, while the NHP and PC tasks operate at the node level, testing the model’s ability to capture local structure by learning different types of entities and relation edges.

\subsection{Training paradigm}
We conducted our experiments using Qwen2.5-7B-Instruct \cite{qwen2025qwen25technicalreport}, Qwen3-8B \cite{yang2025qwen3technicalreport}, and Gemma-7B-IT \cite{gemmateam2024gemmaopenmodelsbased}. The Qwen models are pretrained with the Mixture of Experts (MoE) paradigm, making them efficient and well-suited for complex reasoning tasks such as KG path selection. Qwen3 further introduces explicit “thinking” and “non-thinking” modes to enhance multi-step inference. The Gemma-7B-IT model complements these, offering strong instruction-following and reasoning capabilities, serving as a robust baseline for comparison.

\paragraph{Supervised fine-tuning (SFT)}
For the \textsc{P@10}, \textsc{P@2}, and \textsc{PN@10} tasks, the supervision signal corresponds to identifying the valid paths among the provided candidates. For the NHP and PC tasks, the model is trained to predict either the next node or the next relation edge given the partial path. Training is performed using the cross-entropy loss over the 
task-specific labels. Formally, given an input $x$ with label 
$y \in \{1, \ldots, K\}$, and model output probabilities 
$\hat{p}_\theta(y \mid x)$, the SFT loss is:  
\vspace{-.2em}
\begin{equation} \label{sftLoss}
\mathcal{L}_{\text{SFT}}(\theta) = - \sum_{i=1}^{N} 
\log \hat{p}_\theta(y_i \mid x_i)
\end{equation}

The objective of the SFT stage is to enable the model to internalize the structure of the UMLS KG by learning how different nodes are connected.

\paragraph{DPO}
Proposed by \cite{rafailov2024directpreferenceoptimizationlanguage}, Direct Preference Optimization (DPO) method streamlines the reward modeling and tuning phases typically required in RLHF approaches. In our setting, we apply DPO to models that have already undergone SFT. Due to the maximum likelihood objective of DPO (equation \ref{eq:DPOEq}), the model is encouraged to assign higher probability to valid paths relative to invalid ones. We theorize that by framing our task as a preference-based problem, the fine-tuned models can be further aligned to reliably select positive candidate paths.
\begin{equation}
\label{eq:DPOEq}
\mathcal{L}_{DPO}(\pi_{\theta};\pi_{ref}) = -\mathbb{E}_{(x,y_w,y_l) \sim D}[\log \sigma ( \beta \log \frac{\pi_\theta (y_w | x)}{\pi_{ref} (y_w | x)} - \beta \log \frac{\pi_\theta (y_w | x)}{\pi_{ref} (y_w | x)} )]
\end{equation}


\paragraph{GRPO}
In our path judging paradigm, each training instance includes a group of candidate paths (e.g., 10 paths in P@10 or PN@10), with the objective of assigning higher likelihood to valid diagnostic paths. Group Relative Policy Optimization (GRPO) \cite{shao2024deepseekmathpushinglimitsmathematical} is well suited for this structure, as it directly optimizes over entire groups of candidates by reinforcing the relative preference for correct paths across the set. This allows the model to learn richer path selection strategies beyond isolated pairwise ranking. We define a discrete reward function in Equation \ref{eq:rule-based-reward}, which provides a positive signal exclusively when the model produces the correct label.
\begin{equation} \label{eq: grpo}
\begin{aligned}
\mathcal{L}_{\text{GRPO}}(\theta) &= 
\mathbb{E}_{q \sim P(Q), \, \{o_i\}_{i=1}^G \sim \pi_\theta^{\text{old}}(O \mid q)} 
\Bigg[ 
\frac{1}{G} \sum_{i=1}^{G} \frac{1}{|o_i|} \sum_{t=1}^{|o_i|} 
\min \Bigg( \quad \frac{\pi_\theta(o_{i,t} \mid q, o_{i,<t})}{\pi_\theta^{\text{old}}(o_{i,t} \mid q, o_{i,<t})} \hat{A}_{i,t}, \\
& \quad \text{clip}\Big(\frac{\pi_\theta(o_{i,t} \mid q, o_{i,<t})}{\pi_\theta^{\text{old}}(o_{i,t} \mid q, o_{i,<t})}, 1-\epsilon, 1+\epsilon\Big) \hat{A}_{i,t} 
\Bigg) 
\Bigg] \quad - \beta \, D_{\text{KL}}(\pi_\theta \| \pi_{\text{ref}})
\end{aligned}
\end{equation}

\begin{equation}
\mathcal{R}(x, y) =
\begin{cases} 
1, & \text{if the model prediction } \hat{y} = y\\
0, & \text{otherwise.}
\end{cases}
\label{eq:rule-based-reward}
\end{equation}

\paragraph{Model Merging} \label{modelMerge}
This approach combines two models fine-tuned on different tasks to produce a single model that incorporates aspects of both, without requiring additional training.  The combination is achieved by iterating over the parameters of the two models and computing either a simple average \cite{wortsman2022modelsoupsaveragingweights} or a weighted average \cite{matena2022mergingmodelsfisherweightedaveraging} of the weights. Formally, performing a weighted average in our case looks like:
\[
\theta_{\text{merged}} = \lambda \cdot \theta_{1} + (1 - \lambda) \cdot \theta_{2}
\]
We have merged \textsc{P@10} SFT model with \textsc{P@2} and \textsc{PN@10} models. In both cases, $\theta_1$ corresponds to \textsc{P@10} model parameters. Furthermore, we given higher importance to the parameters of \textsc{P@10} by setting $\lambda = 0.7$.
\paragraph{Reasoning Distillation} \label{reasoningDistil}
Recent approaches frame reward model training as a reasoning task, where Chain-of-Thought (CoT) supervision enhances reward signal quality by modeling deeper inference steps \cite{chen2025rmr1rewardmodelingreasoning}. In our reasoning distillation setup, we sampled 2,577 training examples and used GPT-o3-mini (Microsoft Azure, HIPAA-compliant) to generate explanations for why a correct KG path should be preferred over alternatives (from the \textsc{P@10} setting), given a patient note and ground-truth diagnoses. These high-quality traces then supervised the reward model under multiple training paradigms, in strict accordance with the MIMIC Data Use Agreement.  

\paragraph{RM-R1} \citet{chen2025rm} introduced Reasoning Reward Models (\textsc{ReasRMs}) and proposed a reasoning-oriented training framework, \textsc{RM-R1}, which operates in two stages: (1) distillation of reasoning traces, and (2) reinforcement learning with verifiable rewards.

We adopted this framework by first performing SFT on the generated reasoning traces. This stage trains the model to produce reasoning traces and subsequently predict the preferred candidate path. Following the approach in \cite{chen2025rmr1rewardmodelingreasoning}, we then applied GRPO using a rule-based reward function. Formally, this reward function is defined as:
\[
\mathcal{R}(x, j \mid y_a, y_b) =
\begin{cases} 
1, & \text{if } \hat{l} = l \\ 
-1, & \text{otherwise} 
\end{cases}
\]

\paragraph{DogeRM} \citet{lin2024dogermequippingrewardmodels} introduced the Domain Knowledge Merged Reward Model (\textsc{DogeRM}), which merges a general reward model with a task-specific SFT model. We adopted this strategy by merging our reasoning model, fine-tuned under the \textsc{RM-R1} framework, with the model fine-tuned on the \textsc{P@10} task. We follow the same process as our model merging technique with a simple average of the parameters.
\[
\theta_{\text{MERGE}} = 
0.5 \cdot \theta_{\text{SFT}} + 
0.5 \cdot \theta_{\text{RM}}
\]

\paragraph{Distilling Step by Step (DSS)}
\citet{hsieh2023distillingstepbystepoutperforminglarger}proposed the Distilling Step by Step framework, which extends reasoning distillation by treating reasoning generation and path prediction as two distinct tasks. Specifically, at each training step, the model is prompted twice: first to produce a reasoning trace, and then to predict the correct KG path. The reasoning traces obtained from earlier methods serve as supervision for the reasoning task, while path prediction aligns with the \textsc{P@10} setting. During training, losses for both tasks are computed according to Equation~\ref{sftLoss} and combined to optimize the model jointly.
\[
\mathcal{L}_{total} = 0.5 * \mathcal{L}_{\text{path}} + 0.5 * \mathcal{L}_{\text{rationale}}
\]

\section{Results}

This work investigates two central questions: (i) Can LLMs be trained as reward-style models to verify the relevance of KG paths for diagnosis? (ii) Can such learned reasoning transfer to downstream diagnostic tasks, such as summarization or medical QA? 

Tables \ref{tb: tb1} and \ref{tab: tb2} present results addressing the first research question. These tables compare the performance of various training paradigms across different path selection settings, allowing us to identify the most effective training strategies. In particular, Table \ref{tb: tb1} examines the impact of different path selection finetuning configurations for the Qwen2.5-7B-Instruct model. Additionally, each SFT model was evaluated across all other path selection settings to assess its generalization performance.

\begin{table*}[htbp]
    \centering
    \resizebox{\textwidth}{!}{
    \begin{tabular}{lcccccccccc}
    \toprule \toprule
        \multirow{2}{*}{SFT settings} & \multicolumn{2}{c}{Test on \textsc{P@10}} & \multicolumn{2}{c}{Test on \textsc{P@2}} & \multicolumn{2}{c}{Test on \textsc{NHP}} & \multicolumn{2}{c}{Test on \textsc{PC}} & \multicolumn{2}{c}{Test on \textsc{PN@10}} \\ \cmidrule{2-11}
                       & Rouge-1 & Rouge-L & Rouge-1 & Rouge-L & Rouge-1 & Rouge-L & Rouge-1 & Rouge-L & Rouge-1 & Rouge-L \\ \midrule 
        No SFT   & 19.34 & 18.79 & 78.78 & 78.50 & 1.19  & 1.20  & 4.27  & 3.90  & 28.24 & 27.45 \\  
        SFT on \textsc{P@10} & \cellcolor{lightgray}52.93 & \cellcolor{lightgray}52.64 & 33.98 & 33.31 & 6.26  & 6.22  & 6.64  & 6.35  & 43.14 & 42.65 \\ 
        SFT on \textsc{P@2} & 37.59 & 36.69 &  \cellcolor{lightgray}74.84 &  \cellcolor{lightgray}74.59 & 6.52  & 6.33  & 7.17  & 6.74  & 30.39 & 29.64 \\ 
        SFT on \textsc{NHP} & 8.20  & 7.89  & 8.45  & 7.83  & \cellcolor{lightgray}16.14 & \cellcolor{lightgray}16.09 & 14.33 & 14.24 & 11.34 & 10.61 \\ 
        SFT on \textsc{PC} & 45.62 & 44.74 & 69.90 & 69.25 & 8.72  & 8.69  & \cellcolor{lightgray}18.52 & \cellcolor{lightgray}18.35 & 8.51  & 8.12 \\ 
        SFT on \textsc{PN@10} & 12.53 & 12.14 & 73.89 & 73.65 & 1.37  & 1.37  & 9.05  & 8.84  & \cellcolor{lightgray}18.34 & \cellcolor{lightgray}17.93 \\ 
        \bottomrule \bottomrule
    \end{tabular}
    }
    \caption{Rouge-1 and Rouge-L scores for different SFT configurations of Qwen2.5-7B-Instruct across five tasks. Gray-highlighted cells indicate the scores when the training task matches the test task, demonstrating task-specific effectiveness.}
    \label{tb: tb1}
\end{table*}

In Table \ref{tb: tb1}, the highlighted cells correspond to models that were both finetuned and evaluated on the same setting, where strong performance relative to other models would generally be expected. Interestingly, for the \textsc{P@2} and \textsc{PN@10} settings, the finetuned models perform worse than their non-SFT counterparts. Moreover, the \textsc{NHP} and \textsc{PC} settings appear to be the most challenging, exhibiting the lowest overall performance across models.

\begin{table*}[htbp]
    \centering
    \resizebox{0.95\textwidth}{!}{
    \begin{tabular}{lccccccc}
    \toprule \toprule
        \multirow{2}{*}{Training Paradigm} & \multirow{2}{*}{Task} & \multicolumn{2}{c}{Qwen2.5-7B-Instruct} & \multicolumn{2}{c}{Qwen3-8B} & \multicolumn{2}{c}{Gemma-7B-IT} \\
        \cmidrule{3-8}
                                            & & Rouge-1 & Rouge-L & Rouge-1 & Rouge-L & Rouge-1 & Rouge-L \\ 
        \midrule 
        No SFT (baseline) &          & 19.34   & 18.79   & 4.82    & 4.66    & 26.99   & 26.09   \\ \midrule
        \multirow{4}{*}{SFT} & \textsc{P@10}                 & 52.93   & 52.64   & 18.77   & 18.21   & 17.45   & 16.77   \\ 
         & \textsc{P@2}               & 37.59   & 36.69   & \cellcolor{green!15} 53.17   & \cellcolor{green!15} 52.61   & \cellcolor{green!15} 38.68   & \cellcolor{green!15} 37.93   \\ 
         & \textsc{PC}                 & 45.62   & 44.74   & 3.18    & 3.11    & 16.74   & 16.33   \\ 
         & \textsc{PN@10}                 & 12.53   & 12.14   & 7.65    & 7.02    & 23.83   & 22.84   \\ \midrule
         \multirow{2}{*}{SFT Model Merge} & 0.7 \textsc{P@10} + 0.3\textsc{P@2}  &   \cellcolor{green!15}63.16   & \cellcolor{green!15} 62.35     & 4.70 & 4.53        &  23.34       &   22.58    \\ 
         & 0.7 \textsc{P@10} + 0.3\textsc{PN@10} &   \cellcolor{green!15}62.94      & \cellcolor{green!15}61.89        & 4.35        & 4.19        &  17.36       & 16.38      \\

        
        \midrule
        SFT + DPO  &  \multirow{3}{*}{\textsc{P@10}}                   & 15.90   & 15.06   & 29.35   & 28.92   & 18.01   & 17.22 \\ 
        SFT + GRPO &                   &\textbf{85.03}   & \textbf{84.96}   & 20.04   & 19.58   & 17.70   & 17.07   \\ 
        DogeRM     &                     & 40.73   & 39.97   & 5.55    & 5.38    & 19.79   & 18.60   \\ 
        \midrule
        RM-R1 & \multirow{3}{*}{\textsc{P@10}}   & 46.25   & 45.64   & 5.66    & 5.47    & 20.63   & 19.24 \\ 
        DSS &                     & 59.50   & 59.03   & \cellcolor{green!15} 73.67   & \cellcolor{green!15} 72.95   & \cellcolor{green!15} 84.26   & \cellcolor{green!15} 84.22 \\ 
        DSS + GRPO &          & 59.67   & 59.27   & \textbf{77.24}   & \textbf{76.91}   & \textbf{84.43}   &  \textbf{84.41}   \\ 

        \bottomrule \bottomrule
    \end{tabular}
    }
    \caption{Rouge-1 and Rouge-L scores comparing various training techniques across Qwen2.5-7B-Instruct, Qwen3-8B, and Gemma-7B-IT. Bold values indicate the best training paradigm for each model, while green-highlighted cells show the second-best training paradigms.}
    \label{tab: tb2}
\end{table*}

Table \ref{tab: tb2} compares the different training paradigms. The models are evaluated on the test setting of \textsc{P@10}. DSS combined with GRPO consistently achieves the highest Rouge-1 and Rouge-L scores across all models, highlighting its effectiveness for KG path judgment. In contrast, standard SFT and reward-based methods like DPO and DogeRM underperform, with gains from model merging remaining modest. 
Tables \ref{tab: tb3}, \ref{tab: tb4}, and \ref{tb: tb5} address the question of whether our reward models be effectively applied to diagnostic prediction tasks. For this analysis, we selected the top three performing settings for each model from Table \ref{tab: tb2} and evaluated their ability to generate diagnoses. In Table \ref{tab: tb3}, patient records from the ProbSum test set were provided as input, and the models were tasked with predicting the patient’s diagnoses. To provide a benchmark, we also evaluated non-finetuned models as well as models finetuned directly on the diagnosis prediction task.

We observe that the non-finetuned Gemma-7B-IT and Qwen2.5-7B-Instruct models outperform the non-finetuned Qwen3-8B model on the diagnosis prediction task. After finetuning specifically for diagnosis prediction, the performance of all three models becomes comparable. Notably, the KG-based training strategies explored in this study do not yield significant improvements in diagnosis prediction performance.
\begin{table*}[htbp]
    \centering
    \resizebox{0.95\textwidth}{!}{
    \begin{tabular}{lcccccc}
    \toprule \toprule
        &\multicolumn{2}{c}{Qwen2.5-7B-Instruct} & \multicolumn{2}{c}{Qwen3-8B} & \multicolumn{2}{c}{Gemma-7B-IT} \\ 
        \cmidrule(lr){2-3} \cmidrule(lr){4-5} \cmidrule{6-7}
        &Training & Rouge-L & Training & Rouge-L & Training & Rouge-L \\ 
        \cmidrule(lr){1-7} 
        \multirow{2}{*}{Baseline} & No SFT & 21.10 & No SFT & 5.73 & No SFT & 22.50 \\ 
        & SFT & 22.71 & SFT &  22.19 & SFT & 20.74 \\
        \cmidrule(lr){1-7} 
        \multirow{3}{*}{\makecell[l]{Reward-like \\ Reasoning SFT}} & \makecell[c]{SFT Model Merge\\ 0.7 \textsc{P@10} + 0.3 \textsc{P@2}} & 21.70 & SFT \textsc{P@2} & 19.47 & SFT \textsc{P@2} & 15.00 \\ 
        \cmidrule(lr){2-3} \cmidrule(lr){4-5} \cmidrule{6-7}
        & \makecell[c]{SFT Model Merge\\ 0.7 \textsc{P@10} + 0.3 \textsc{PN@10}} &21.60 & \makecell[l]{DSS} & 17.18 & \makecell[c]{DSS} &22.17 \\  
        \cmidrule(lr){2-3} \cmidrule(lr){4-5} \cmidrule{6-7}
        & SFT + GRPO &12.98 & \makecell[c]{DSS + GRPO} &15.93 & \makecell[l]{DSS + GRPO} & 22.17 \\ 
        \bottomrule \bottomrule
    \end{tabular}
    }
    \caption{Rouge-L scores for diagnosis prediction on ProbSum test set. Top three performing training paradigms for each model from \ref{tab: tb2} are evaluated alongside baseline methods (no SFT and SFT for diagnosis prediction task).}
    \label{tab: tb3}
\end{table*}

Results on the MedQA dataset are summarized in Table \ref{tab: tb4}. For this evaluation, we also trained task-specific models as baseline. Notably, for Qwen2.5-7B and Qwen3-8B, our training strategies show modest improvements over the baseline models. 

\begin{table*}[htbp]
    \centering
    \begin{tabular}{lccc}
    \toprule \toprule
        Training & Qwen2.5-7B-Instruct & Qwen3-8B & Gemma-7B-IT \\ \midrule 
        No SFT               & 63.74 & 44.22 & 31.08 \\
        SFT on MedQA                    & 63.34 & 50.59 & 50.59 \\
        \makecell[l]{Best training paradigm\\ from the checkpoints in Table \ref{tab: tb3}}   & 64.90 & 56.97 & 43.42 \\
        \bottomrule \bottomrule
    \end{tabular}
    \caption{QA results on MedQA dataset. Best performing training paradigm for each model from \ref{tab: tb3} are evaluated alongside baseline methods (no SFT and SFT on MedQA).}
    \label{tab: tb4}
\end{table*}


\begin{table*}[htbp]
    \centering
    \resizebox{0.99\textwidth}{!}{
    \begin{tabular}{lcccccc}
    \toprule \toprule
        \multirow{3}{*}{PDSQI-9 Metrics} & \multicolumn{3}{c}{Baseline (no SFT)} & \multicolumn{3}{c}{Best output from Table \ref{tab: tb3} per LLM} \\
        \cmidrule(lr){2-4}\cmidrule(lr){5-7}
                                        & Qwen2.5-7B-Instruct & Qwen3-8B & Gemma-7B-IT & Qwen2.5-7B-Instruct & Qwen3-8B & Gemma-7B-IT \\
                                        &         &           &          & \makecell[c]{(SFT Model Merge\\ 0.7 \textsc{P@10} + 0.3 \textsc{P@2})}  & (SFT \textsc{P@2})     & (DSS) \\ \midrule
        Accuracy Extractive             & 1.62    & 2.04      & 1.70     & 1.76 $\uparrow$   & 1.46      & 1.54      \\
        Thoroughness                    & 1.94    & 1.68      & 1.84     & 2.00 $\uparrow$   & 1.44      & 1.50      \\
        Usefulness                      & 2.48    & 2.20      & 2.24     & 2.44              & 1.64      & 1.94      \\
        Organization                    & 3.68    & 2.30      & 3.32     & 3.80 $\uparrow$   & 1.84      & 2.70      \\
        Comprehensibility               & 4.54    & 3.28      & 4.50     & 4.58 $\uparrow$   & 2.84      & 4.64 $\uparrow$     \\
        Succinctness                    & 2.70    & 2.10      & 3.10     & 2.78 $\uparrow$   & 1.38      & 3.92 $\uparrow$     \\
        Synthesis Abstraction           & 1.80    & 2.18      & 1.60     & 2.18 $\uparrow$   & 1.20      & 0.54      \\
        \midrule
        Average Score          & 2.68 & 2.25 & 2.61 & 2.79 $\uparrow$ & 1.68 & 2.39 \\
        \bottomrule \bottomrule
    \end{tabular}
    }
    
    \caption{ProbSum results using PDSQI-9 criteria with GPT-o4-mini as the judge.}
    \label{tb: tb5}
\end{table*}

Table \ref{tb: tb5} presents the PDSQI-9 scores for the best-performing models identified in Table \ref{tab: tb3}, alongside non-finetuned models as a baseline. For this evaluation, patient records from the ProbSum test set were provided as input, and the models were tasked with generating both the diagnosis and the corresponding reasoning. Even in this setting, we observe inconsistencies among the finetuned models. Specifically, Qwen2.5-7B-Instruct shows only marginal improvements across the metrics, while Qwen3-8B and Gemma-7B-IT exhibit decreases in most metrics relative to their baselines.

Collectively, Tables \ref{tab: tb3}, \ref{tab: tb4}, and \ref{tb: tb5} indicate that proficiency in identifying valid KG paths does not necessarily translate to improved performance on downstream tasks such as diagnostic reasoning.

\section{Discussion: Brittleness and Promise}

Our findings highlight both the brittleness and the emerging promise of fine tuning LLMs on KG reasoning tasks. Across a series of SFT experiments, we observe that models often perform well on the format they are trained on, yet struggle to generalize to even closely related reasoning formats. This inconsistency reveals a persistent challenge in aligning LLMs with the structured, multi hop logic of KG based clinical reasoning.

\paragraph{Brittleness: shallow adaptation and fragile generalization} 
Table~\ref{tb: tb1} shows that models fine tuned on one task (e.g., \textsc{P@10} or \textsc{NHP}) exhibit strong performance on that specific test but often perform worse than the no SFT baseline on other reasoning tasks. This suggests that SFT primarily teaches models to conform to a particular task format, rather than imparting deeper reasoning capabilities over the KG.






\begin{figure}[ht]
    \centering
    \begin{subfigure}[t]{0.48\textwidth}
        \centering
        \tiny 
        \begin{minted}[frame=single, bgcolor=gray!10]{text}
Prompt: 
...............Patient Progress Notes ...............
Given the PATH 'Elevated k->has_member|Chronic kidney 
disease (smq)->member_of', predict the next hop of 
the PATH

Model: |Hyperkalemia
Actual: |K excess
        \end{minted}
        \vspace{-.15in}
        \caption{Qwen2.5-7B-Instruct SFT on \textsc{NHP}}
        \label{sample:nhp}
    \end{subfigure}
    \hfill
    \begin{subfigure}[t]{0.48\textwidth}
        \centering
        \tiny 
        \begin{minted}[frame=single, bgcolor=gray!10]{text}
Prompt: 
..............Patient Progress Notes ...............
Given the PATH 'Elevated k', complete the rest 
of the PATH.

Model: ->subset_includes_item
Actual: ->has_member|Chronic kidney disease (smq)
        ->member_of|K excess
        \end{minted}
        \vspace{-.15in}
        \caption{Qwen2.5-7B-Instruct SFT on \textsc{PC}}
        \label{sample:pc}
    \end{subfigure}
    \vspace{-.1in}
    \caption{Sample responses from Qwen2.5-7B-Instruct for NHP and PC tasks.}
    \label{fig:samples}
\end{figure}

Error analysis (Figure~\ref{fig:samples}) further underscores this limitation. Models frequently hallucinate incorrect relations, terminate paths prematurely, or output synonyms (e.g., ``Hyperkalemia'') instead of canonical KG terms (e.g., ``K excess''), breaking alignment with ground truth outputs. In other cases, models generate relations that are semantically equivalent to the labeled ones but differ in surface form. For example, predicting `$\rightarrow$subset\_includes\_item` instead of the labeled `$\rightarrow$has\_member`. These examples illustrate a key limitation of our current evaluation functions: they rely on exact string matching and do not account for synonymy or ontological equivalence in the KG. Future evaluation strategies should incorporate semantic type alignment and relation-level equivalence to better reflect clinically valid reasoning.

We further observe that models fine tuned on \textsc{NHP} (Next Hop Prediction) struggle when transferred to \textsc{P@10} tasks, particularly when a clear starting node is missing. In such cases, the model often begins its path from an arbitrary node, resulting in broken or incomplete reasoning chains.

\paragraph{Promise: very specific reward optimization and checkpoint merging} 
Despite this brittleness, we also find signals of promise. As shown in Table~\ref{tab: tb2} and \ref{tb: tb5}, the Qwen2.5-7B-Instruct model fine tuned using our GRPO method outperforms all other variants, including baselines and other distillation setups, on human aligned PDSQI-9 criteria. Improvements are seen not only in surface qualities like organization and succinctness, but also in deeper aspects such as extractive accuracy and abstract synthesis. This suggests that when the reward signal is well aligned and the model is guided by end task utility, it can meaningfully integrate KG derived reasoning into its generative output. Interestingly, this promise is model specific: only one LLM (Qwen2.5-7B) showed consistent gains across multiple criteria. Others (e.g., Gemma-7b-it) performed well on select metrics (e.g., succinctness) but failed on others, highlighting that model architecture and initialization likely affect how reasoning priors interact with KG based supervision. 

Overall, our results point to a narrow path forward: while current SFT approaches do not reliably induce generalizable reasoning over knowledge graphs, carefully aligned reward based tuning (e.g., GRPO) holds potential for unlocking clinically useful capabilities. It is also important to note that our experiments use a single knowledge graph (UMLS) as the underlying structure. While other KGs may offer different reasoning affordances, UMLS is one of the most widely adopted in clinical applications, and our findings are therefore likely to reflect meaningful patterns in KG based diagnostic reasoning more broadly. 

To move beyond brittle adaptations, future work must focus on strategies that emphasize structural reasoning, canonical grounding, and consistency across task formats. These may include curriculum learning, constrained decoding over KG structures, and richer reward modeling that integrates both symbolic accuracy and language fluency. As LLMs continue to evolve, the interplay between symbolic supervision and generative capacity offers a promising, though still fragile, frontier.

\section{Conclusion}
Framing reasoning as a reward-like judgment task, we introduced a suite of path-based training setups and evaluated their effects on both KG verification and downstream diagnostic tasks. While certain strategies, such as GRPO and CoT distillation, enhance the model’s ability to assess KG path validity, their benefits do not consistently generalize to broader clinical reasoning settings. Our findings reveal a narrow corridor of promise: reward supervision over KGs can encode reasoning signals, but realizing their full clinical utility will require deeper integration of structure, context, and generative capabilities. 






\bibliographystyle{unsrtnat}
\bibliography{custom}

\end{document}